\documentclass[runningheads]{llncs}

\usepackage[T1]{fontenc}
\usepackage[american]{babel}
\addto\captionsamerican{%
  \renewcommand{\abstractname}{Abstract.}%
  \renewcommand{\figurename}{Fig.}%
}
\usepackage{amsmath,amssymb,amsfonts}
\usepackage{algorithm}
\usepackage{algorithmic}
\usepackage{booktabs}
\usepackage{colortbl}
\usepackage{graphicx}
\usepackage{multirow}
\usepackage{pifont}
\usepackage{siunitx}
\usepackage{textcomp}
\usepackage[table]{xcolor}
\usepackage{xurl}
\usepackage[hidelinks]{hyperref}
\usepackage[sort,compress]{cite}
\usepackage{marvosym}

\newcommand{\citep}[1]{\cite{#1}}
\newcommand{\parencite}[1]{\cite{#1}}
\newcommand{\textcite}[1]{\cite{#1}}

\title{Replaying the Movie in Your Mind: Decoding Dynamic Visual Perception and Recall from EEG}
\titlerunning{Decoding Dynamic Visual Perception and Recall from EEG}

\author{Zikai Fang\thanks{These authors contributed equally to this work.} \and
Tian-Yi Zhou\protect\footnotemark[1] \and
Xuan-Hao Liu \and Yixi Li \and Xiaoyu Xing \and
Yutong Pei \and Bao-Liang Lu \and
Wei-Long Zheng\thanks{Corresponding author.}}
\authorrunning{Z. Fang et al.}

\institute{School of Computer Science, Shanghai Jiao Tong University, Shanghai, China\\
\email{\{sky\_walker,ty\_zhou,weilong\}@sjtu.edu.cn} 
}

\begin{document}

\maketitle

\begin{abstract}
Decoding dynamic visual content from EEG has advanced rapidly for visual perception, yet it remains unclear which aspects of neural representations are preserved when people internally replay dynamic scenes. We introduce PAIR, an EEG benchmark collected from 20 participants using an interleaved, clip-wise paradigm that pairs naturalistic video viewing with immediate visual recall. This design enables a comparison between externally driven perception and internally generated recall for the same stimulus content within a fixed trial sequence. We benchmark multiple decoding models on hierarchical semantic classification at both fine-grained and coarse-grained levels, meta-feature decoding tasks covering motion, color, object number, face, and human presence, and a state-decoding task that discriminates perception from recall. Within-subject validation results show numerically similar performance for several model--task combinations, alongside higher perception accuracy in two coarse-grained comparisons after multiple-comparison correction. In contrast, perception and recall are highly separable at the state level under the present validation protocol. State accuracy remains high after alpha-band ablation, although this does not exclude ocular or other task-related contributions.

\end{abstract}

\section{Introduction}
\label{sec:intro}

Visual brain decoding, often regarded as the scientific realization of "mind-reading", seeks to interpret the neural patterns to infer what a person is seeing from their brain signals and how the human brain processes visual stimuli\cite{du2018reconstructing}. This field has witnessed a paradigm shift from decoding simple, static categories\cite{fang2020reconstructing, naselaris2009bayesian_reconstruction} to reconstructing complex, dynamic visual experiences \cite{li2024visual, gong2024neuroclips}. Early foundational works primarily focused on reconstructing static images~\citep{zhang2025cognitioncapturer, scotti2023reconstructing}. However, natural vision is not a sequence of frozen frames but a continuous stream of temporal events. Addressing this challenge, recent research has pivoted toward video reconstruction, seeking to decode continuous visual streams from brain signals~\cite{lu2025animateyourthoughts, huang2025mindev, zhou2026mindcine}. This evolution has been fueled by the release of large-scale public datasets. In the fMRI domain, datasets such as CC2017 have established rigorous standards for spatial reconstruction~\citep{wen2018neural}. In parallel, leveraging millisecond-level temporal resolution, the EEG community has introduced benchmarks like SEED-DV, which are pivotal for decoding the rapid temporal dynamics inherent in visual processing~\citep{liu2024eeg2video}.

While existing datasets have propelled the decoding of visual perception, they predominantly characterize neural responses elicited while subjects passively view video stimuli, largely overlooking an equally fundamental cognitive mechanism: visual imagery. Often described as the ability to "see with the mind's eye," imagery is central to understanding how the brain simulates reality~\citep{pearson2015mental, kosslyn2001neural}. A pivotal yet unresolved question in cognitive science is the extent of the neural overlap between these two states---specifically, whether internally generated imagery reuses the same representational codes as externally driven perception or relies on distinct reconstruction processes~\citep{dijkstra2019shared}. Although this perception-imagery relationship has been scrutinized using static stimuli~\citep{reddy2010reading}, the mechanisms underlying dynamic naturalistic imagery remain obscure. Unlike static images, dynamic mental replay necessitates the reconstruction of temporal structure---motion and event evolution---in addition to semantic content~\citep{dijkstra2018temporal}. Crucially, strictly controlled benchmarks that directly compare perception and imagery for the \emph{same} dynamic stimuli---under aligned temporal conditions and task demands---are notably absent, limiting our ability to probe these shared or distinct neural dynamics.

To bridge this gap, we introduce an EEG benchmark, called \textbf{P}erception \textbf{A}nd \textbf{I}magery \textbf{R}ecall (PAIR), collected from 20 healthy subjects engaged in a controlled interleaved perception--recall paradigm. Unlike passive viewing datasets, our protocol requires participants to watch naturalistic videos and immediately mentally replay the visual content, pairing EEG recordings of external perception and internal imagery for each clip. Rather than assuming a particular direction of effect~\citep{hochstein2002view, dijkstra2019shared}, we use this benchmark to compare semantic and meta-feature decoding during dynamic recall and perception, with matched stimulus content and a fixed trial sequence.


In conclusion, our contributions can be summarized as follows:
\begin{enumerate}
  \item We introduce an EEG dataset for dynamic video viewing and immediate imagery recall. The clip-wise paradigm pairs the two conditions for the same content within a fixed trial sequence.
  \item We benchmark traditional and deep models on 20- and 4-class semantic decoding, meta-feature classification, and state decoding.
  \item Our within-subject validation results show high perception--recall state separability, while content-decoding differences depend on the task and model.
\end{enumerate}

\section{Related Work}

\subsection{Perception--Imagery Overlap and Mental Replay}
A core debate in cognitive science concerns the extent to which visual imagery relies on the same representational substrate as perception.
A synthesis of recent neuroimaging work suggests substantial overlap between perception and imagery in visual, parietal, and frontal regions, with imagery characterized by stronger top-down influences and the absence of bottom-up sensory drive~\citep{dijkstra2019shared}.
Complementing fMRI evidence, EEG studies have begun to isolate oscillatory signatures of shared content representations; For example, Dijkstra et al.~\cite{dijkstra2018temporal} reported that cross-generalizable representations between perceived and imagined objects emerge prominently in the alpha band over posterior sensors.
These lines of work motivate paradigms that compare perception and imagery under tightly matched stimulus content and timing, especially for naturalistic \emph{dynamic} scenes.

\subsection{EEG Decoding of Visual Content and Naturalistic Stimuli}
Decoding visual information from EEG has progressed from controlled, static stimuli to more complex and naturalistic settings, supported by increasingly large datasets and improved modeling~\citep{liu2025mindcross, huang2025mindev}.
Early demonstrations showed that EEG contains discriminative information about visual categories, enabling brain-driven visual classification \parencite{spampinato2017deep}.
More recent work has emphasized richer stimulus sets and temporal representational dynamics \parencite{grootswagers2019representational,gifford2022large,li2025brainflora}.
Together, these studies establish EEG as a practical modality for probing time-resolved representational content during perception and provide methodological precedents for decoding semantic structure from continuous or rapidly changing visual input.

\subsection{Deep Learning Models for EEG Decoding}

End-to-end deep learning has revolutionized EEG decoding, with architectures ranging from deep ConvNets~\citep{schirrmeister2017deep} and compact CNNs like EEGNet~\citep{lawhern2018eegnet} to hybrid CNN-Transformer models~\citep{song2023eegconformer} now serving as standard backbones for spatiotemporal feature extraction. However, while these models have achieved state-of-the-art performance in perceptual tasks, strictly controlled benchmarks that leverage them to directly compare neural representations between \textit{continuous} video perception and \textit{dynamic} mental replay remain notably absent. In this study, we utilize these advanced spatiotemporal architectures to bridge this gap, systematically evaluating the fidelity of semantic and meta-feature preservation across these distinct cognitive states.
\section{GCN-LSTM Baseline}

\subsection{Spatio-Temporal Graph Neural Network (GCN-LSTM)}

To effectively capture both the spatial topology of electrode distribution and the long-term temporal dependencies of EEG signals, we employ a hybrid Spatio-Temporal GCN-LSTM model. While this architecture has been successfully applied to EEG tasks such as seizure detection \cite{covert2019temporal} and emotion recognition \cite{zhang2019spatial}, we evaluate it here as a baseline for \textbf{Dynamic Visual Recall} classification.

\begin{figure}[htbp]
  \centering
  \includegraphics[width=\linewidth]{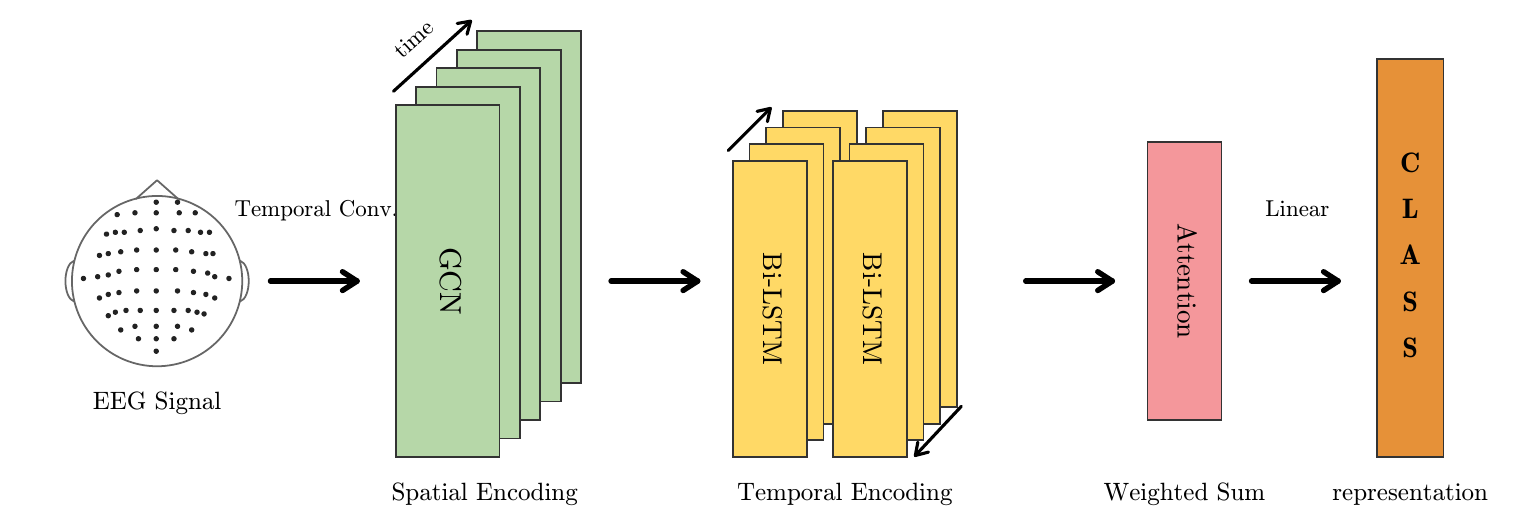}
  \caption{Model structure of GCN-LSTM.}
  \label{fig:GCN_LSTM}
\end{figure}

As illustrated in Fig.\ref{fig:GCN_LSTM}, the architecture integrates graph convolutions with recurrent units. Let the input EEG segment be denoted as $X \in \mathbb{R}^{C \times T}$, where $C$ represents the number of electrode channels and $T$ denotes the time samples. Notably, this model was exclusively benchmarked on raw EEG data; we did not utilize PSD or DE features because their low dimensionality (spanning only 5 frequency bands) makes them unsuitable for the fine-grained temporal modeling capabilities of the LSTM.

\textbf{Graph Construction.} We model the spatial relationships between electrodes as a graph $\mathcal{G} = (\mathcal{V}, \mathcal{E})$. The nodes $\mathcal{V}$ correspond to the $C$ EEG electrodes. The adjacency matrix $A \in \mathbb{R}^{C \times C}$, representing the edges $\mathcal{E}$, is constructed based on the physical geometry of the sensors. Specifically, the edge weight between electrode $i$ and $j$ is defined by the inverse square of their Euclidean distance, calculated as $A_{ij} \propto \frac{1}{dist(i, j)^2}$. We apply a threshold 0.5 to $A$ to ensure a sparse representation of significant spatial connections.

\textbf{Spatio-Temporal Encoding.} The model has three stages:
\begin{enumerate}
    \item \textbf{Shallow Temporal Feature Extraction:} The raw input $X$ is first processed by a temporal convolution layer to extract short-term temporal patterns and reduce the sampling frequency, yielding an intermediate representation.
    
    \item \textbf{Spatial GCN Module:} To capture spatial dependencies, we treat the features at each time step as graph signals. We employ Graph Convolutional Networks (GCN) to aggregate information from neighboring nodes. The layer-wise propagation rule is given by:
    \begin{equation}
        H^{(l+1)} = \sigma(\tilde{D}^{-\frac{1}{2}}\tilde{A}\tilde{D}^{-\frac{1}{2}} H^{(l)} W^{(l)}),
    \end{equation}
    where $\tilde{A} = A + I$ is the adjacency matrix with self-connections, $\tilde{D}$ is the degree matrix, and $W^{(l)}$ represents trainable weights.
    
    \item \textbf{Temporal LSTM with Attention:} The sequence of spatially encoded features is then flattened and fed into a Bi-directional Long Short-Term Memory (Bi-LSTM) network to model global temporal dynamics. Finally, an attention mechanism aggregates the LSTM hidden states by assigning adaptive weights to different time steps, generating a context vector for the final classification task.
\end{enumerate}

\section{Dataset and Benchmark}
\subsection{Participants}
Data were collected from 20 healthy subjects (10 males; 10 females). All participants had normal or corrected-to-normal vision and provided written informed consent.

\begin{figure}[htbp]
  \centering
  \includegraphics[width=0.8\linewidth]{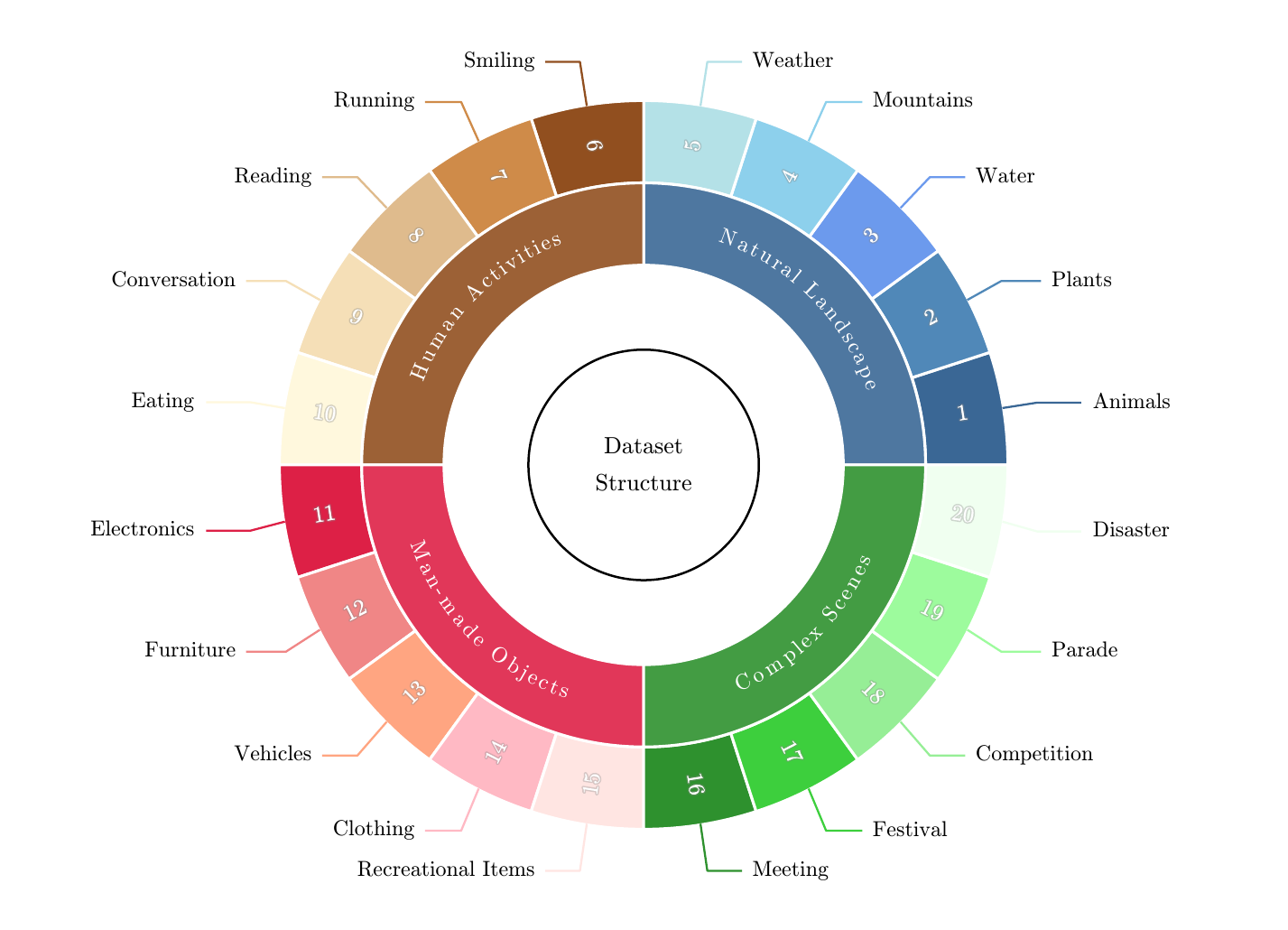}
  \caption{The hierarchical structure of the dataset, consisting of 4 major categories and 20 subcategories.}
  \label{fig:categories}
\end{figure}

\subsection{Stimuli and Experimental Paradigm}

\subsubsection{Stimuli Construction.} 
We constructed a custom video dataset sourced from public platforms (YouTube and Bilibili). The dataset comprises 250 short video clips (2 seconds each), hierarchically organized into four major categories: \textit{Natural Landscape}, \textit{Human Activities}, \textit{Man-made Objects}, and \textit{Complex Scenes}. As illustrated in Fig.~\ref{fig:categories}, these major categories branch into 20 fine-grained subcategories (e.g., Animals, Smiling, Electronics, and Meeting), covering a diverse range of semantic concepts.

\subsubsection{Experimental Protocol.} 
\begin{figure}[htbp]
  \centering
  \includegraphics[width=\linewidth]{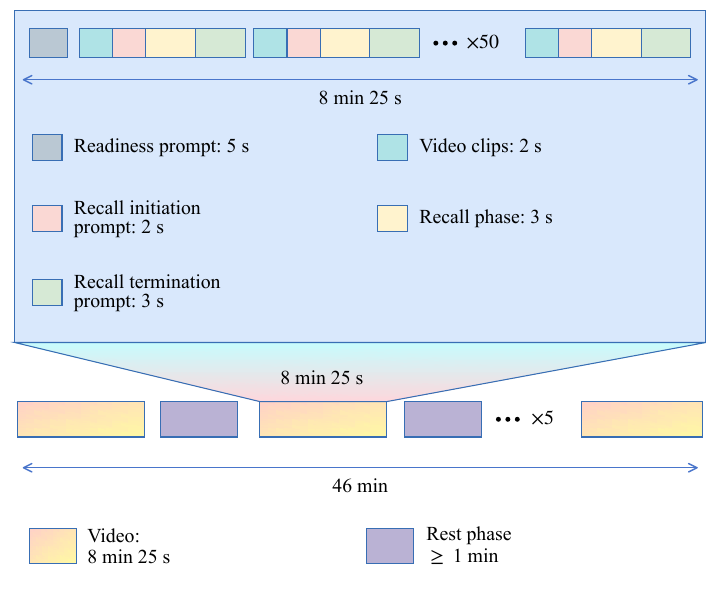}
  \caption{The experimental paradigm. Each block contains 50 trials with a prepare-watch-recall structure.}
  \label{fig:paradigm}
\end{figure}

Data were collected from 20 healthy subjects. Each subject participated in 3 sessions. During each session, subjects viewed 5 continuous video blocks. As shown in Fig.~\ref{fig:paradigm}, each block consists of 50 distinct trials (video clips). 

The temporal structure of a single block is designed as follows: A block begins with a 5-second readiness prompt. Each trial then proceeds with a 2-second video clip presentation, followed by a 2-second recall initiation prompt, a 3-second recall phase, and a 3-second termination prompt (serving as a break before the next trial). This cycle repeats 50 times, resulting in a total duration of 505 seconds (approx. 8 min 25 s) per video block. 
\subsubsection{Data Acquisition.}
To ensure the quality of the acquired EEG data, the experiments were conducted in a controlled laboratory environment to minimize noise and other environmental disturbances. 62-channel EEG signals were collected by an active AgCl electrode cap with an international 10-10 system. The EEG signals were acquired using the ESI NeuroScan System at a sampling rate of 1000 Hz. In addition to EEG, we used a Tobii Pro Fusion eye tracker to record eye movements at 250~Hz. The present study analyzed EEG only; eye-tracking recordings were not included in the reported analyses.

\subsection{Data Preprocessing}
\label{sec:preprocessing}
Raw EEG signals were band-pass filtered (0.1--100 Hz) and processed using Independent Component Analysis (ICA) to attenuate ocular and muscular artifacts. The EEG signals were downsampled to 200~Hz to reduce computational cost. The continuous recordings were then segmented into distinct epochs corresponding to the Perception and Recall conditions. Content decoding used the full 2-s perception and 3-s recall epochs. For raw-EEG state decoding, recall was cropped to 0.5--2.5~s, giving 2-s inputs for both conditions. Each input was standardized using its own mean and standard deviation; content-training inputs additionally received Gaussian noise with standard deviation 0.05.

\subsection{Baselines and Implementation Details}

\subsubsection{Baselines.} We first computed Power Spectral Density (PSD) and Differential Entropy (DE) across five standard frequency bands ($\delta$: 1--4~Hz, $\theta$: 4--8~Hz, $\alpha$: 8--14~Hz, $\beta$: 14--31~Hz, and $\gamma$: 31--99~Hz) using the Short-Time Fourier Transform (STFT) with a Hanning window.

To validate the effectiveness of our approach, we benchmark it against several state-of-the-art decoding models, including ShallowNet, DeepNet, EEGNet, Conformer, and TSConv. Notably, we also include \textbf{GLFNet} proposed by~\citep{liu2024eeg2video} as a key baseline, as it effectively integrates global and local EEG features and has demonstrated robust performance in dynamic visual tasks.

\subsubsection{Implementation Details.}
Neural networks used Adam with learning rates of $10^{-3}$ or $3\times10^{-4}$, weight decay $10^{-4}$, and early stopping with patience 40 (up to 400 epochs). Batch sizes were 16 for content decoding and 128 for state decoding. SVM used an RBF kernel with $C=1$; content-decoding SVM results are reported for raw EEG only.

Models were fitted separately for each participant and recording session. For content decoding, the 250 epochs in each session were randomly divided into 200 training and 50 validation epochs. State decoding used the first 200 epochs from each condition for training and the remaining 50 for validation, preserving clip pairing across partitions. Neural-network scores are the best validation accuracies selected during early stopping; SVM scores are validation accuracies for the fixed configuration. No independent test set was used.

For reporting, the three session scores were averaged within each participant, followed by the mean and sample standard deviation across 20 participants. Exploratory two-sided paired $t$-tests compared perception and recall for all 84 model--feature--task combinations retained in Tables~\ref{tab:Perception} and~\ref{tab:Recall}, with Holm correction across these comparisons. Mean paired differences are accompanied by unadjusted 95\% confidence intervals. These analyses describe the archived validation scores and do not correct selection bias.

\subsection{Classification Tasks}
\label{sec:classification_tasks}

To comprehensively evaluate the decoding performance, we adopted the benchmark suite established by EEG-VP~\citep{liu2024eeg2video} and extended it with State Decoding, a task unique to our dataset. This resulting suite comprises three distinct classification tasks: Semantic Decoding, Meta-feature Decoding, and State Decoding. Together, they range from high-level conceptual understanding to low-level visual and recall feature retrieval and mental state discrimination.

\subsubsection{Task 1: Semantic Decoding}
This task evaluates the model's ability to distinguish specific semantic concepts from neural activity. We defined the ground truth labels at two hierarchical levels of granularity: Fine-grained Classification (20-class), Coarse-grained Classification (4-class)

Both classification levels were applied to the \textit{Perception} and \textit{Recall} conditions independently to compare the semantic distinctiveness across mental states.

\subsubsection{Task 2: Meta-feature Decoding}
Beyond semantic identity, we decoded mid-level visual and cognitive attributes. These "meta-features" serve as a probe to investigate whether specific physical properties of the stimuli are preserved in the neural representations. The ground truth labels were generated as follows:

\begin{itemize}
    \item \textbf{Motion Speed (Fast/slow):} We quantified the motion information of each video stimulus using the Optical Flow Score (OFS). The videos were then split into binary classes (\textit{Fast/slow}) using the implemented threshold of 0.6427.
    
    \item \textbf{Dominant Color:} We extracted the dominant color features from each video and categorized them into five \textit{atmosphere-oriented} classes: \textit{Neutral Light}, \textit{Earth \& Dark}, \textit{Cool Tones}, \textit{Green Nature}, and \textit{Warm Vibrant}.
    
    \item \textbf{Biological/Social Attributes:} We manually annotated the presence of specific visual elements. 
    \begin{itemize}
        \item \textit{Human:} Binary label indicating the presence of a human figure.
        \item \textit{Face:} Binary label indicating the presence of a visible face.
        \item \textit{Number:} 3-class label indicating the number of distinct objects in the scene:
        \begin{equation}
            y = \mathbb{I}(\text{count} > 1) + \mathbb{I}(\text{count} > 4).
        \end{equation}
    \end{itemize}
\end{itemize}

The chance rows in the tables indicate uniform random guessing. Class frequencies are not generally balanced: across the 250 stimuli, majority-class proportions are 60.4\% for Face, 54.4\% for Human, and 43.6\% for Number. These stimulus-set proportions are descriptive, not validation-set baseline scores.

\subsubsection{Task 3: State Decoding}
This task aims to distinguish the subject's underlying cognitive state. We formulated this as a binary classification problem:
\begin{equation}
    y_{state} = 
    \begin{cases} 
    0 & \text{if } x \in \mathcal{D}_{perception} \\
    1 & \text{if } x \in \mathcal{D}_{recall}
    \end{cases},
\end{equation}
where $x$ represents the EEG sample.

\section{Results}

\subsection{Benchmarking Decoding Models}
\label{sec:benchmark}

In this section, we report validation performance for decoding dynamic video content from EEG signals. We compared traditional machine learning approaches (e.g., SVM) with deep learning architectures, including DeepNet, Conformer, and GCN-LSTM. The classification accuracies across different tasks are summarized in Tables~\ref{tab:Perception} and~\ref{tab:Recall}.

\begin{table*}[h]
\centering
\caption{Perception validation accuracy (\%; mean $\pm$ SD across 20 participants after averaging three sessions each). Neural-network scores are best validation accuracies; SVM uses its fixed configuration. Highest means within each feature group are in \textbf{bold}.}
\label{tab:Perception}
\footnotesize
\renewcommand{\arraystretch}{1} 
\resizebox{\textwidth}{!}{%
\begin{tabular}{lccccccc}
\toprule
\textbf{Model} & \textbf{20-c} & \textbf{4-c} & \textbf{Fast/slow} & \textbf{Color} & \textbf{Number} & \textbf{Face} & \textbf{Human} \\
\midrule
Uniform chance & 5 & 25 & 50 & 20 & 33.3 & 50 & 50 \\
\midrule
\multicolumn{8}{c}{Raw EEG Signals} \\
\midrule
ShallowNet & $10.50 \pm 1.67$ & $38.27 \pm 3.63$ & $60.90 \pm 3.50$ & $27.60 \pm 2.66$ & $45.50 \pm 3.72$ & $62.40 \pm 3.05$ & $59.93 \pm 3.27$ \\
DeepNet & $11.10 \pm 1.23$ & $38.87 \pm 3.39$ & $62.20 \pm 2.45$ & $29.73 \pm 2.76$ & $46.43 \pm 2.77$ & $62.93 \pm 2.92$ & $61.03 \pm 4.15$ \\
EEGNet & $10.47 \pm 1.42$ & $37.53 \pm 3.41$ & $61.53 \pm 3.19$ & $29.40 \pm 3.42$ & $45.93 \pm 3.31$ & $63.00 \pm 3.01$ & $60.27 \pm 3.50$ \\
Conformer & $10.20 \pm 1.48$ & $37.17 \pm 3.48$ & $\mathbf{62.47 \pm 4.33}$ & $29.60 \pm 3.40$ & $\mathbf{49.13 \pm 2.75}$ & $\mathbf{65.20 \pm 3.73}$ & $\mathbf{62.33 \pm 3.18}$ \\
TSConv & $10.43 \pm 1.99$ & $36.37 \pm 2.54$ & $60.27 \pm 3.30$ & $29.10 \pm 3.25$ & $45.53 \pm 2.85$ & $62.67 \pm 2.80$ & $59.37 \pm 3.16$ \\
GLFNet & $\mathbf{11.57 \pm 2.30}$ & $\mathbf{39.53 \pm 3.30}$ & $61.30 \pm 2.77$ & $28.87 \pm 2.69$ & $46.87 \pm 3.62$ & $62.47 \pm 2.63$ & $61.40 \pm 3.01$ \\
SVM & $4.23 \pm 1.67$ & $27.40 \pm 2.92$ & $49.70 \pm 4.58$ & $20.47 \pm 3.09$ & $36.43 \pm 2.77$ & $53.40 \pm 3.15$ & $51.17 \pm 4.57$ \\
GCN-LSTM & $10.70 \pm 1.72$ & $36.17 \pm 2.33$ & $61.20 \pm 2.90$ & $\mathbf{30.77 \pm 3.06}$ & $48.63 \pm 2.75$ & $64.97 \pm 2.47$ & $61.30 \pm 2.84$ \\
\midrule
\multicolumn{8}{c}{PSD Features} \\
\midrule
\textbf{Model} & \textbf{20-c} & \textbf{4-c} & \textbf{Fast/slow} & \textbf{Color} & \textbf{Number} & \textbf{Face} & \textbf{Human} \\
\midrule
MLPNet & $8.13 \pm 1.52$ & $35.20 \pm 2.51$ & $57.47 \pm 3.37$ & $28.23 \pm 3.33$ & $46.73 \pm 3.33$ & $63.10 \pm 2.26$ & $58.93 \pm 3.21$ \\
GLFNet-MLP & $\mathbf{8.80 \pm 1.89}$ & $\mathbf{35.80 \pm 3.32}$ & $\mathbf{60.30 \pm 2.89}$ & $\mathbf{29.13 \pm 2.53}$ & $\mathbf{47.13 \pm 3.43}$ & $\mathbf{63.57 \pm 2.45}$ & $\mathbf{59.20 \pm 3.36}$ \\
\midrule
\multicolumn{8}{c}{DE Features} \\
\midrule
\textbf{Model} & \textbf{20-c} & \textbf{4-c} & \textbf{Fast/slow} & \textbf{Color} & \textbf{Number} & \textbf{Face} & \textbf{Human} \\
\midrule
MLPNet & $10.87 \pm 1.73$ & $\mathbf{36.87 \pm 2.50}$ & $61.00 \pm 2.34$ & $29.43 \pm 2.82$ & $48.10 \pm 2.80$ & $64.73 \pm 2.45$ & $60.00 \pm 2.88$ \\
GLFNet-MLP & $\mathbf{11.27 \pm 1.66}$ & $36.73 \pm 2.18$ & $\mathbf{61.73 \pm 3.23}$ & $\mathbf{29.83 \pm 2.30}$ & $\mathbf{48.63 \pm 2.41}$ & $\mathbf{65.20 \pm 2.63}$ & $\mathbf{60.63 \pm 3.24}$ \\
\bottomrule

\end{tabular}%
}
\end{table*}

\begin{table*}[!h]
\centering
\caption{Recall validation accuracy (\%; mean $\pm$ SD across 20 participants). Aggregation and score selection follow Table~\ref{tab:Perception}. Highest means within each feature group are in \textbf{bold}.}
\label{tab:Recall}
\footnotesize
\renewcommand{\arraystretch}{1} 
\resizebox{\textwidth}{!}{%
\begin{tabular}{lccccccc}
\toprule
\textbf{Model} & \textbf{20-c} & \textbf{4-c} & \textbf{Fast/slow} & \textbf{Color} & \textbf{Number} & \textbf{Face} & \textbf{Human} \\
\midrule
Uniform chance & 5 & 25 & 50 & 20 & 33.3 & 50 & 50 \\
\midrule
\multicolumn{8}{c}{Raw EEG Signals (Recall)} \\
\midrule
ShallowNet & $9.40 \pm 1.67$ & $33.97 \pm 3.23$ & $61.03 \pm 3.27$ & $27.70 \pm 2.57$ & $44.70 \pm 3.28$ & $62.83 \pm 3.16$ & $60.20 \pm 3.73$ \\
DeepNet & $9.80 \pm 2.13$ & $34.47 \pm 2.81$ & $\mathbf{62.43 \pm 4.15}$ & $28.97 \pm 2.64$ & $46.20 \pm 3.71$ & $63.40 \pm 3.32$ & $60.93 \pm 2.69$ \\
EEGNet & $9.63 \pm 1.58$ & $34.10 \pm 3.30$ & $59.43 \pm 3.32$ & $28.13 \pm 2.97$ & $45.67 \pm 2.64$ & $62.10 \pm 4.05$ & $60.40 \pm 2.66$ \\
Conformer & $10.20 \pm 1.69$ & $35.83 \pm 2.31$ & $61.63 \pm 2.89$ & $29.30 \pm 2.09$ & $47.83 \pm 3.11$ & $64.63 \pm 3.09$ & $\mathbf{61.53 \pm 3.25}$ \\
TSConv & $9.23 \pm 1.58$ & $33.97 \pm 3.00$ & $61.90 \pm 3.18$ & $28.10 \pm 2.74$ & $44.17 \pm 3.63$ & $61.57 \pm 2.60$ & $59.80 \pm 3.01$ \\
GLFNet & $9.17 \pm 1.45$ & $33.67 \pm 2.38$ & $60.90 \pm 3.14$ & $27.97 \pm 3.20$ & $45.03 \pm 3.89$ & $63.43 \pm 3.44$ & $60.43 \pm 3.58$ \\
SVM & $3.83 \pm 1.42$ & $23.03 \pm 3.01$ & $48.23 \pm 3.41$ & $18.97 \pm 2.81$ & $36.43 \pm 4.02$ & $52.17 \pm 5.11$ & $50.93 \pm 1.83$ \\
GCN-LSTM & $\mathbf{10.73 \pm 1.43}$ & $\mathbf{36.53 \pm 2.87}$ & $62.33 \pm 2.56$ & $\mathbf{30.07 \pm 2.32}$ & $\mathbf{48.60 \pm 3.22}$ & $\mathbf{64.93 \pm 2.89}$ & $60.80 \pm 2.81$ \\
\midrule
\multicolumn{8}{c}{PSD Features} \\
\midrule
\textbf{Model} & \textbf{20-c} & \textbf{4-c} & \textbf{Fast/slow} & \textbf{Color} & \textbf{Number} & \textbf{Face} & \textbf{Human} \\
\midrule
MLPNet & $7.47 \pm 1.68$ & $33.33 \pm 3.27$ & $56.37 \pm 4.14$ & $26.33 \pm 2.50$ & $46.10 \pm 3.55$ & $\mathbf{62.80 \pm 2.68}$ & $\mathbf{57.23 \pm 2.52}$ \\
GLFNet-MLP & $\mathbf{7.80 \pm 1.52}$ & $\mathbf{34.00 \pm 2.79}$ & $\mathbf{59.47 \pm 2.95}$ & $\mathbf{28.17 \pm 2.63}$ & $\mathbf{46.30 \pm 4.31}$ & $\mathbf{62.80 \pm 2.67}$ & $57.03 \pm 2.82$ \\
\midrule
\multicolumn{8}{c}{DE Features} \\
\midrule
\textbf{Model} & \textbf{20-c} & \textbf{4-c} & \textbf{Fast/slow} & \textbf{Color} & \textbf{Number} & \textbf{Face} & \textbf{Human} \\
\midrule
MLPNet & $\mathbf{10.00 \pm 1.16}$ & $\mathbf{35.00 \pm 2.31}$ & $60.57 \pm 2.72$ & $29.20 \pm 2.50$ & $47.63 \pm 3.69$ & $64.50 \pm 1.67$ & $\mathbf{61.50 \pm 2.63}$ \\
GLFNet-MLP & $9.90 \pm 1.47$ & $34.60 \pm 2.18$ & $\mathbf{60.77 \pm 3.28}$ & $\mathbf{29.27 \pm 2.55}$ & $\mathbf{47.73 \pm 3.18}$ & $\mathbf{64.53 \pm 1.60}$ & $60.90 \pm 2.63$ \\
\bottomrule
\end{tabular}}
\end{table*}

\subsubsection{Comparison of Decoding Models}
As shown in Tables~\ref{tab:Perception} and~\ref{tab:Recall}, raw-EEG neural networks yielded higher recorded validation scores than SVM. This is a descriptive comparison: neural networks were selected using validation accuracy, whereas SVM used a fixed configuration. The scores should also be interpreted in light of class imbalance rather than uniform chance alone.

\subsubsection{Performance of Spatiotemporal Models}
A closer inspection of the results shows that the \textbf{GCN-LSTM} model achieved the highest mean raw-EEG validation accuracy in several \textit{Recall} tasks, including 4-c, color, number, and face classification. These observations support its inclusion as a spatiotemporal baseline for PAIR, without establishing a mechanism-specific advantage.

\subsection{Semantic and Meta-feature Decoding Analysis}

\begin{figure}[htbp]
  \centering
  \includegraphics[width=\linewidth]{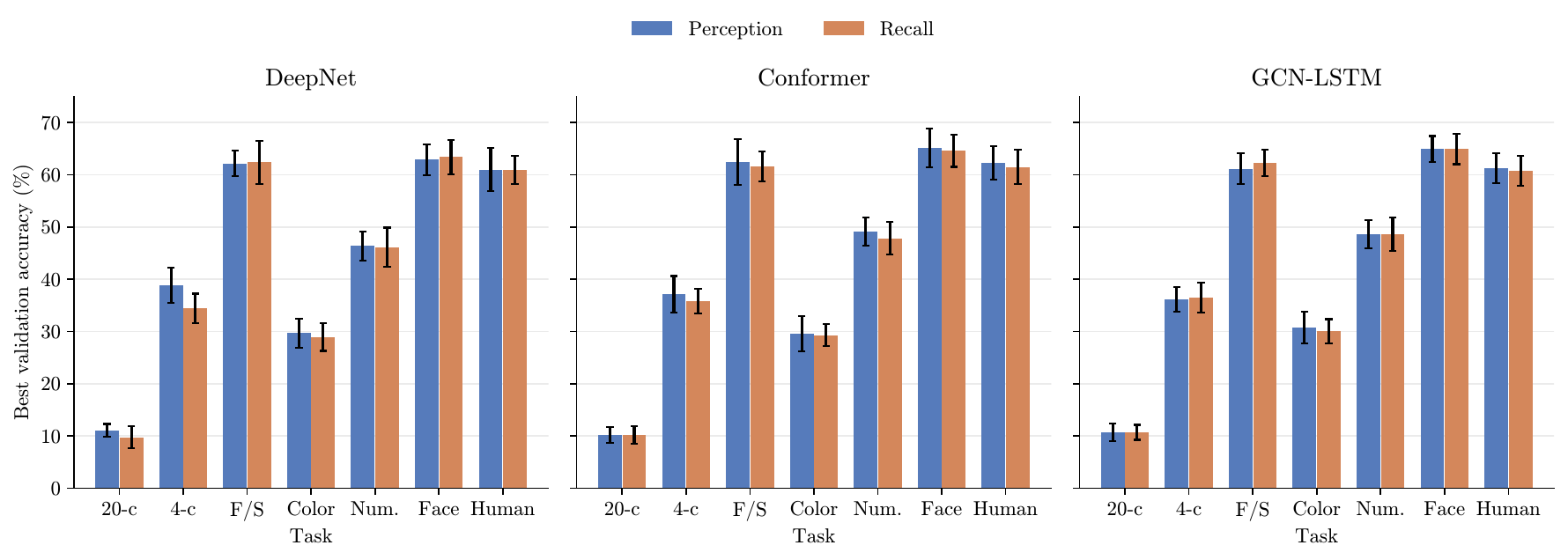}
  \caption{Perception (blue) and recall (orange) across semantic and meta-feature tasks. Bars show mean best validation accuracy; error bars show SD across 20 participants after averaging sessions. Similar bars do not establish statistical equivalence.}
    \label{fig:perception_recall}
\end{figure}

To compare content decoding across conditions, we evaluated coarse-grained (4-class) and fine-grained (20-class) semantics together with meta-features. Fig.~\ref{fig:perception_recall} illustrates performance for three representative models.

Several model--task combinations yielded numerically similar scores, but differences were not uniform. After Holm correction across 84 comparisons, perception exceeded recall for raw-EEG 4-class decoding with DeepNet (mean paired difference 4.40 percentage points, 95\% CI [2.65, 6.15], adjusted $p=0.0038$) and GLFNet (5.87 points, 95\% CI [4.22, 7.51], adjusted $p<0.001$). The remaining comparisons did not reach the corrected threshold of 0.05. In particular, GCN-LSTM yielded close mean scores across conditions; nonsignificant differences do not establish equivalent performance or shared representations. All comparisons remain exploratory because they use validation-selected scores.

\subsection{Analysis of Perception vs. Recall State Decoding}

Table~\ref{tab:PerceptionOrRecall} shows high validation accuracy for discriminating visual perception and recall. Differential Entropy (DE) features yielded the highest mean scores under this protocol. These results describe separability between the experimental conditions and do not isolate imagery-specific neural activity.

\begin{table}[!t]
    \centering
    \caption{State validation accuracy (\%; mean $\pm$ SD across 20 participants), with full-band inputs and alpha ablation. Score selection follows Table~\ref{tab:Perception}. $\Delta$ is the full-minus-ablated mean difference in percentage points, computed before rounding.}
    \label{tab:PerceptionOrRecall}
    \renewcommand{\arraystretch}{1.05}
    \setlength{\tabcolsep}{6pt}
    \begin{tabular}{llccc}
        \toprule
        \textbf{Feature} & \textbf{Method} & \textbf{Full Band} & \textbf{w/o Alpha} & \textbf{$\Delta$} \\
        \midrule
        -- & Chance Level & 50.00 & 50.00 & 0.00 \\
        \midrule
        \multirow{8}{*}{Raw EEG}
        & DeepNet & $84.77 \pm 7.87$ & $79.23 \pm 10.22$ & 5.53 \\
        & SVM & $88.10 \pm 4.80$ & $86.73 \pm 5.11$ & 1.37 \\
        & EEGNet & $94.88 \pm 3.39$ & $92.53 \pm 3.84$ & 2.35 \\
        & Conformer & $95.07 \pm 3.48$ & $93.88 \pm 3.69$ & 1.18 \\
        & GCN-LSTM & $95.52 \pm 3.33$ & $95.57 \pm 1.76$ & -0.05 \\
        & ShallowNet & $95.97 \pm 3.35$ & $94.78 \pm 2.84$ & 1.18 \\
        & GLFNet & $96.52 \pm 2.97$ & $95.22 \pm 3.25$ & 1.30 \\
        & TSConv & $\mathbf{96.53 \pm 2.15}$ & $\mathbf{96.08 \pm 2.18}$ & 0.45 \\
        \midrule
        \multirow{3}{*}{PSD}
        & SVM & $83.43 \pm 7.78$ & $81.52 \pm 7.36$ & 1.92 \\
        & MLPNet & $95.37 \pm 3.59$ & $91.70 \pm 4.50$ & 3.67 \\
        & GLFNet-MLP & $\mathbf{95.48 \pm 3.93}$ & $\mathbf{93.23 \pm 4.21}$ & 2.25 \\
        \midrule
        \multirow{3}{*}{DE}
        & SVM & $95.12 \pm 2.79$ & $94.52 \pm 3.15$ & 0.60 \\
        & MLPNet & $98.57 \pm 1.18$ & $98.20 \pm 1.59$ & 0.37 \\
        & GLFNet-MLP & $\mathbf{98.58 \pm 1.17}$ & $\mathbf{98.33 \pm 1.55}$ & 0.25 \\
        \bottomrule
    \end{tabular}
\end{table}

A potential contributor to state discrimination is alpha activity. The archived ablation runs used an 8--13~Hz band-stop filter for raw EEG and omitted the 8--14~Hz alpha feature for PSD/DE before model fitting. Thus, the raw and feature-based ablations have slightly different frequency definitions. As shown in Table~\ref{tab:PerceptionOrRecall}, TSConv decreased by 0.45 percentage points and GLFNet-MLP on DE features by 0.25 points. High validation accuracy persisted after these ablations, indicating that the removed alpha information was not necessary for high scores in these models. This does not exclude other spectral, ocular, or task-related contributions.

\begin{figure}[htbp]
  \centering
  \includegraphics[width=\linewidth]{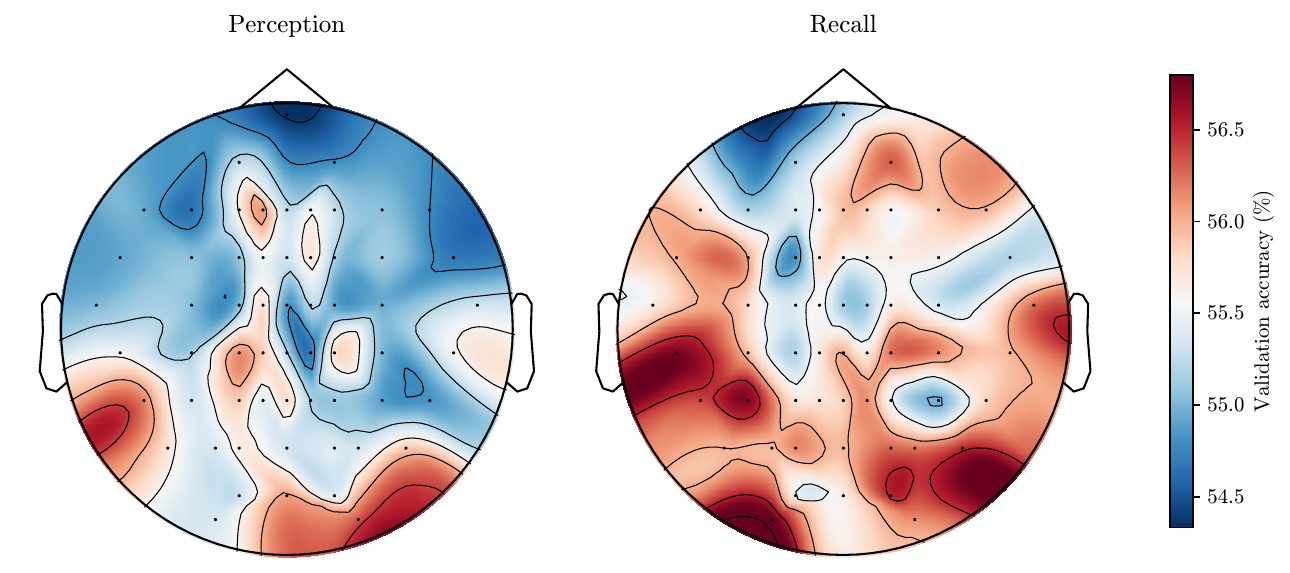}
  \caption{Mean sensor-level validation accuracy for the `Human' meta-feature during perception and recall, using a shared single-channel Conformer and a common color scale. Values are averaged across sessions and participants; no spatial significance is implied.}
  \label{fig:topomap}
\end{figure}

To illustrate spatial variation, Fig.~\ref{fig:topomap} shows validation accuracy for the `Human' meta-feature at each sensor. A shared single-channel Conformer was fitted across electrodes, with epoch selection based on their mean validation accuracy. Perception and recall show different sensor-level distributions, including variation across anterior and posterior locations. The maps are descriptive: they neither establish statistically significant spatial differences nor identify cortical sources or frontal executive-network recruitment.

\section{Conclusion}
Our benchmarking results provide initial within-subject validation baselines for dynamic visual perception and recall, with GCN-LSTM yielding the highest mean accuracy on several recall tasks.

Several models yielded numerically similar validation accuracies for perception and recall, although the differences varied across tasks and models. The paired analysis identified higher perception accuracy for two coarse-grained comparisons after correction. Similar decoding scores alone do not establish equivalent performance or shared neural representations.

State decoding yielded high validation accuracy, including after alpha-band ablation. The topographic analysis illustrates differences across anterior and posterior sensors, without identifying cortical sources. PAIR provides a basis for further investigation of content decoding and state differences under more stringent evaluation and artifact controls.

\paragraph{Limitations.}
Validation data were used for both epoch selection and performance reporting, so neural-network scores may be optimistically biased and model comparisons are exploratory. The 20-participant, within-subject protocol does not establish cross-subject generalization. Content epochs also differ in duration between perception and recall. Although ICA attenuated artifacts, eye-tracking control analyses were not performed; residual ocular activity, visual input, and task demands may contribute to state separability. Sensor-level maps cannot establish recruitment of specific cortical networks.

\paragraph{Data and code availability.}
Benchmark code, stimulus labels, analysis materials, and the anonymized EEG dataset are available at \url{https://github.com/freesky3/PAIR} and the \href{https://huggingface.co/datasets/skywalker-p/PAIR/tree/v1.0.0}{PAIR v1.0.0 dataset}. The dataset is released under CC BY-NC 4.0; eye-tracking recordings are not included.

\begin{credits}
\subsubsection{\ackname}
\begingroup
\emergencystretch=1em
This work was supported in part by grants from Brain Science and Brain-like Intelligence Technology-National Science and Technology Major Project (2025ZD0218900), National Natural Science Foundation of China (62376158), STI 2030-Major Projects+2022ZD0208500, National Key Research and Development Program of China (2024YFC3606800), Medical-Engineering Interdisciplinary Research Foundation of Shanghai Jiao Tong University ``Jiao Tong Star'' Program (YG2023ZD25, YG2024ZD25 and YG2024QNA03), Shanghai Jiao Tong University 2030 Initiative, the Lingang Laboratory (Grant No. LGL-1987), GuangCi Professorship Program of RuiJin Hospital Shanghai Jiao Tong University School of Medicine,\linebreak[2] and Shanghai Jiao Tong University SCS-Shanghai Emotionhelper Technology Co., Ltd Joint Laboratory of Affective Brain-Computer Interfaces.
\par\endgroup
\subsubsection{\discintname}
The authors have no competing interests to declare that are relevant to the content of this article.
\end{credits}

\bibliographystyle{splncs04}
\bibliography{ref}

\end{document}